\documentclass[10pt,twocolumn,letterpaper]{article}

\usepackage{iccv}              

\definecolor{iccvblue}{rgb}{0.21,0.49,0.74}
\usepackage[pagebackref,breaklinks,colorlinks,allcolors=iccvblue]{hyperref}

\usepackage{url}
\usepackage{graphicx}
\usepackage{booktabs}
\usepackage{multirow}
\usepackage{subcaption}
\usepackage{siunitx}
\usepackage{wrapfig}
\usepackage{pifont}

\usepackage{cuted}

\def\paperID{3856} 
\def\confName{ICCV}
\def\confYear{2025}

\title{SynFER: Towards Boosting Facial Expression Recognition with Synthetic Data}

\author{
Xilin He$^{1,9*}$, Cheng Luo$^{4,5*}$, Xiaole Xian$^{1}$, Bing Li$^{4}$, Muhammad Haris Khan$^{8}$ \\Zongyuan Ge$^{5}$, Weicheng Xie$^{1,2,3\dagger}$, Siyang Song$^{6\dagger}$, Linlin Shen$^{7\dagger}$, Bernard Ghanem$^{4}$, Xiangyu Yue$^{9}$\\
$^1$School of Computer Science \& Software Engineering, Shenzhen University, China\\
$^2$Guangdong Laboratory of Artificial Intelligence and Digital Economy (SZ), Shenzhen, China\\
$^3$Guangdong Provincial Key Laboratory of Intelligent Information Processing, Shenzhen University, China\\
$^4$KAUST, $^5$ Monash University, $^6$School of Computer Science, University of Exeter, UK \\
$^7$ Computer Vision Institute, School of Artificial Intelligence, Shenzhen University, China \\$^8$ MBZUAI $^9$ The Chinese University of Hong Kong\\
}

\begin{document}
\maketitle
\begin{abstract}
Facial expression datasets remain limited in scale due to the subjectivity of annotations and the labor-intensive nature of data collection. This limitation poses a significant challenge for developing modern deep learning-based facial expression analysis models, particularly foundation models, that rely on large-scale data for optimal performance. To tackle the overarching and complex challenge, instead of introducing a new large-scale dataset, we introduce SynFER (Synthesis of Facial Expressions with Refined Control), a novel synthetic framework for synthesizing facial expression image data based on high-level textual descriptions as well as more fine-grained and precise control through facial action units.
To ensure the quality and reliability of the synthetic data, we propose a semantic guidance technique to steer the generation process and a pseudo-label generator to help rectify the facial expression labels for the synthetic images.
To demonstrate the generation fidelity and the effectiveness of the synthetic data from SynFER, we conduct extensive experiments on representation learning using both synthetic data and real-world data. Results validate the efficacy of our approach and the synthetic data. 
Notably, our approach achieves a 67.23\% classification accuracy on AffectNet when training solely with synthetic data equivalent to the AffectNet training set size, which increases to 69.84\% when scaling up to five times the original size. Code is available \href{https://github.com/C0notSilly/SynFER}{here}.
\vspace{-1em}
\end{abstract}
\renewcommand{\thefootnote}{}
\footnotetext{* These authors contributed equally.}
\footnotetext{$\dagger$ indicates corresponding authors.}
    
\section{Introduction}
\label{sec:intro}

Facial Expression Recognition (FER) is at the forefront of advancing AI’s ability to interpret human emotions, opening new frontiers for various human-centered applications. From automatic emotion detection to early interventions in mental health \cite{ringeval2019avec}, accurate pain assessment \cite{huang2024naturalistic}, and enhancing human-computer interaction \cite{abdat2011human}, the potential impact of FER systems is profound \cite{moin2023emotion,sajjad2023comprehensive,zhu2023toward}. 
In recent years, learning-based FER models have gained significant traction due to their promising performances \cite{li2020deep,zhang2021relative,farzaneh2021facial}. However, despite recent advancements in network architectures and learning methodologies, the progress of existing FER models has been hindered by the inadequate scale and quality of available training data, underscoring the need to expand datasets with high-quality data to push the boundaries of FER capabilities.

Existing FER datasets, such as CK+ (953 sequences) \cite{dataset_ckplus}, FER-2013 (30,000 48$\times$48 images) \cite{dataset_ferplus}, RAF-DB (29,672 images) \cite{dataset_rafdb}, AFEW (113,355 images) \cite{dhall2017individual}, and SFEW (1,766 images) \cite{dataset_sfew}, 
are small compared to popular image datasets for general image processing (e.g., ImageNet \cite{deng2009imagenet} with 1.4 million images and Laion \cite{schuhmann2022laion} with billion-level data). 
While AffectNet \cite{dataset_affectnet} compiles a large number of facial images from the web, 
it still suffers from vital drawbacks. A considerable portion of AffectNet’s images are low-quality, and its annotations often contain incorrect labels, which impairs the training process of FER models \cite{Le_2023_WACV, icme_label_noise}.
Consequently, the absence of high-quality and large-scale FER datasets has delayed the development of FER foundation models. However, collecting a large-scale FER dataset with high-quality facial images and meticulous annotations is almost an unrealistic endeavor due to substantial financial and time costs, ethical concerns around facial data collection, and limited resources for large-scale acquisition. Additionally, the subjective interpretation of facial expressions results in inconsistent labeling by annotators, which exacerbates variability and hinders the creation of reliable datasets.

\begin{figure*}[!htp]
    \centering
    \includegraphics[width=0.9\textwidth]{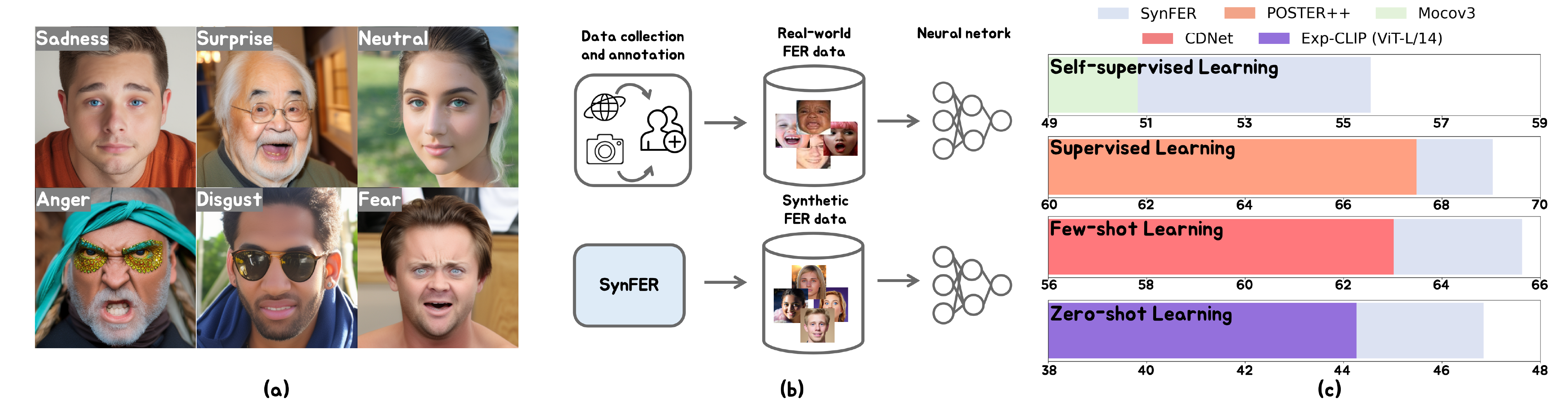} 
    \vspace{-3mm}
        \caption{(a) Examples of synthetic facial expression data generated by our SynFER model, (b) Comparison of training paradigms: training with real-world data versus training with synthetic facial expression data and (c) Performance boost from SynFER generated Data in supervised, self-supervised, zero-shot, and few-shot (5-shot) learning tasks.}
    \label{fig:vis}
    \vspace{-1em}
\end{figure*}

To address the challenges in developing FER models, we turn to explore the generation paradigm to synthesize high-quality facial expression images paired with reliable labels. This approach draws inspiration from successful strategies employed to expand annotated datasets for other computer vision tasks, such as semantic segmentation \cite{baranchuk2021label, chen2019learning, li2022bigdatasetgan} and depth estimation \cite{atapour2018real, cheng2020s, guizilini2022learning}.
These advances leverage powerful generative models such as Stable Diffusion \cite{ldm} and DALL-E \cite{Dalle}, which capture intricate natural image patterns. By tapping into these models, researchers have generated realistic images with their corresponding annotations, thereby boosting model performance.
However, applying diffusion models to synthesize facial expression images with reliable FER labels presents two major challenges. 
(1) the training sets used by these generative models often lack diverse facial expression data, limiting their ability to produce images that capture subtle and nuanced emotional semantics; 
and (2) prior approaches to generate annotations for synthetic images focused on tangible attributes such as pixel-wise layouts, or depth maps. In contrast, facial expressions convey abstract and subjective emotions, making the generation of precise and reliable expression labels much more complex. 
To the best of our knowledge, none of the existing methods can simultaneously conduct fine-grained control for facial expression generation and generate robust categorical facial expression labels leveraging diffusion-based features.




In this paper, instead of a dataset, we present SynFER, a leading data synthesis pipeline capable of synthesizing unlimited and realistic facial expression images paired with reliable expression labels, to drive advancements in FER models.
To address the shortcomings of existing FER datasets, which often lack expression-related text paired with facial images, we introduce FEText, a unique hybrid dataset created by curating and filtering data from existing FER and high-quality face datasets. This vision-language dataset serves as the foundation for training our generative model to synthesize facial expression data.
To ensure fine-grained control and faithful generation of facial expression images, we inject facial action unit (FAU) information and semantic guidance from external pre-trained FER models. Building upon this, we propose FERAnno, the first diffusion-based label calibrator for FER, which automatically generates reliable annotations for the synthesized images. 
Together, these innovations position SynFER as a powerful tool for producing large-scale, high-quality facial expression data, offering a significant resource for the development of FER models.

We investigate the effectiveness of the synthetic data across various learning paradigms, demonstrating consistent and modest improvement in model performance. 
As shown in Fig. \ref{fig:vis}(c), training with the synthetic data yields significant performance boosts across various learning paradigms.
Notably, pre-training on the synthetic data (Fig. \ref{fig:vis}(c)) improves self-supervised learning model MoCov3 \cite{chen2021empirical} on AffectNet, surpassing real-world data pre-training. In supervised learning, SynFER improves accuracy by +1.55\% for the state-of-the-art FER model, POSTER++ \cite{mao2024poster++}, on AffectNet.
We further explore the performance scaling of the synthetic data, revealing further gains as dataset size increases.
Our key contributions are:
\begin{itemize}
    \item We introduce FEText, the first dataset of facial expression-related image-text pairs, providing a crucial resource for advancing FER tasks.
    \item We propose SynFER, the first diffusion-based data synthesis pipeline for FER, integrating FAUs information and semantic guidance to achieve fine-grained control and faithful expression generation. Additionally, FERAnno, a novel diffusion-based label calibrator, is designed to automatically refine and enhance the annotations of synthesized facial expression images.
    \item Extensive experiments across six datasets and four learning paradigms demonstrate the effectiveness of the proposed SynFER, validating the quality and scalability of its synthesized output.
\end{itemize}
\section{Related Work}

\textbf{Facial Expression Recognition (FER):} 
Recent success in deep learning (DL) has largely boosted the performance of the FER task, despite the substantial data requirements for training DL models. To address the limited training data in FER, previous methods mainly focus on developing different learning paradigms, including semi-supervised learning \cite{adacm,Yu_2023_CVPR,cho2024rmfer}, transfer learning \cite{fer_li2022deep, ruan2022adaptive} and multi-task learning \cite{liu2023uncertain,li2023compound}. 
For example, Ada-CM \cite{adacm} learns a confidence margin to make full use of the unlabeled facial expression data in a semi-supervised manner.
Despite achieving performance gains for FER, these methods remain constrained by limited data. Recently, researchers have explored an alternative data-driven perspective of introducing large-scale face datasets from other facial analysis tasks (e.g., face recognition \cite{face2exp}). Meta-Face2Exp \cite{face2exp} utilizes large-scale face recognition data to enhance FER by matching the feature distribution between face recognition and FER. However, face data drawn from these datasets lack diverse facial expressions, and thereby couldn't fully unlock the potential of large-scale data in FER.



\noindent\textbf{Synthetic Data:} 
Recently, growing attention has been paid to the advanced generative models (e.g., Generative Adversarial Networks (GANs) \cite{goodfellow2020generative} and Diffusion Models \cite{rombach2022high}), which are typically flexible to synthesize training images for a wider range of downstream tasks, including classification \cite{frid2018synthetic, azizi2023synthetic}, face recognition \cite{kim2023dcface, boutros2023idiff}, semantic segmentation \cite{datasetdiffusion, wu2024datasetdm, wu2023diffumask} and human pose estimation \cite{feng2023diffpose, zhou2023diff3dhpe}. In particular, some studies pioneer the capabilities of powerful pre-trained diffusion generative models on natural images \cite{datasetdiffusion, wu2024datasetdm, li2023grounded}.
For example, DatasetDM \cite{wu2024datasetdm} further introduces a generalized perception decoder to parse the rich latent space of the pre-trained diffusion model for various downstream tasks. 
Despite the growing adoption of diffusion models in synthetic data generation, their application to anatomically grounded facial expression synthesis remains critically underexplored. 
While recent diffusion-based methods like FineFace \cite{fineface} and EmojiDiff \cite{jiang2024emojidiff} achieve visually compelling facial expression synthesis through AU-guided editing or reference-driven generation, their utility as synthetic training data for FER remains unverified. However, these methods prioritize perceptual quality over functional utility, lacking explicit mechanisms to align generated expressions with FER label semantics or to validate downstream task performance. In this paper, we attempt to investigate the potential and feasibility of synthetic images for FER tasks.
\section{Preliminaries}

Diffusion models include a forward process that adds Gaussian noise $\epsilon$ to convert a clean sample $x_0$ to noise sample $x_T$, and a backward process that iteratively performs denoising from $x_T$ to $x_0$, where $T$ represents the total number of timesteps. The forward process of injecting noise is:
\begin{small}
    \begin{equation}
    \label{eq::add_noise}
    x_t = \sqrt{\alpha_t}x_0 + \sqrt{1-\alpha_t}\epsilon
\end{equation}
\end{small}
$x_t$ is the noise feature at timestep $t$ and $\alpha_t$ is a predetermined hyperparam. for sampling $x_t$ with a given noise scheduler \cite{song2020denoising}.
In backward process of denoising, given input noise $x_t$ sampled from a Gaussian distribution, a learnable network $\epsilon_{\theta}$ estimates the noise at each timestep $t$ with condition $c$. $x_{t-1}$, the feature at the previous timestep is:
\begin{small}
\begin{equation}
    \label{eq::diff_step}
    x_{t-1} = \frac{\sqrt{\alpha_{t-1}}}{\sqrt{\alpha_{t}}} x_t + \sqrt{\alpha_{t-1}} (\sqrt{\frac{1}{\alpha_{t-1}} - 1} - \sqrt{\frac{1}{\alpha_t} -1}) \epsilon_{\theta}(x_t, t, c)
\end{equation}
\end{small}
During training, the noise estimation network $\epsilon_{\theta}$ is guided to conduct denoising with condition $c$ by the learning objective:
\begin{small}
\begin{equation}
    \label{eq::diff_loss}
    \min_{\theta} \mathbb{E}_{x_0,\epsilon\sim\mathcal{N}(\mathbf{0},\mathbf{I}),c,t}\|\epsilon-\epsilon_\theta(x_t,c,t)\|_2^2,
\end{equation}  
\end{small}
With its powerful capability to model complex data distributions, the diffusion model serves as the foundation for generating high-quality FER data. Our SynFER framework is the pioneering work that explores the use of diffusion models to synthesize affective modalities.









\section{Methodology}

We first introduce i) the overall synthetic pipeline for generating facial expression image-label pairs. Next, we detail ii) our approach for producing high-fidelity facial expression images, which are controlled through high-level text descriptions (Sec.\ref{sec:fetext}), fine-grained facial action units corresponding to localized facial muscles (Sec.\ref{sec:au}), and a semantic guidance technique (Sec.\ref{sec:sg}). Finally, we introduce iii) the FER annotation crafter (FERAnno), a crucial component that thoroughly understands the synthetic facial expression data and automatically generates accurate annotations accordingly (Sec.\ref{sec:anno}). This pipeline ensures both precision and reliability in facial expression generation and labeling.

\begin{figure*}
    \centering
    \includegraphics[width=0.9\textwidth]{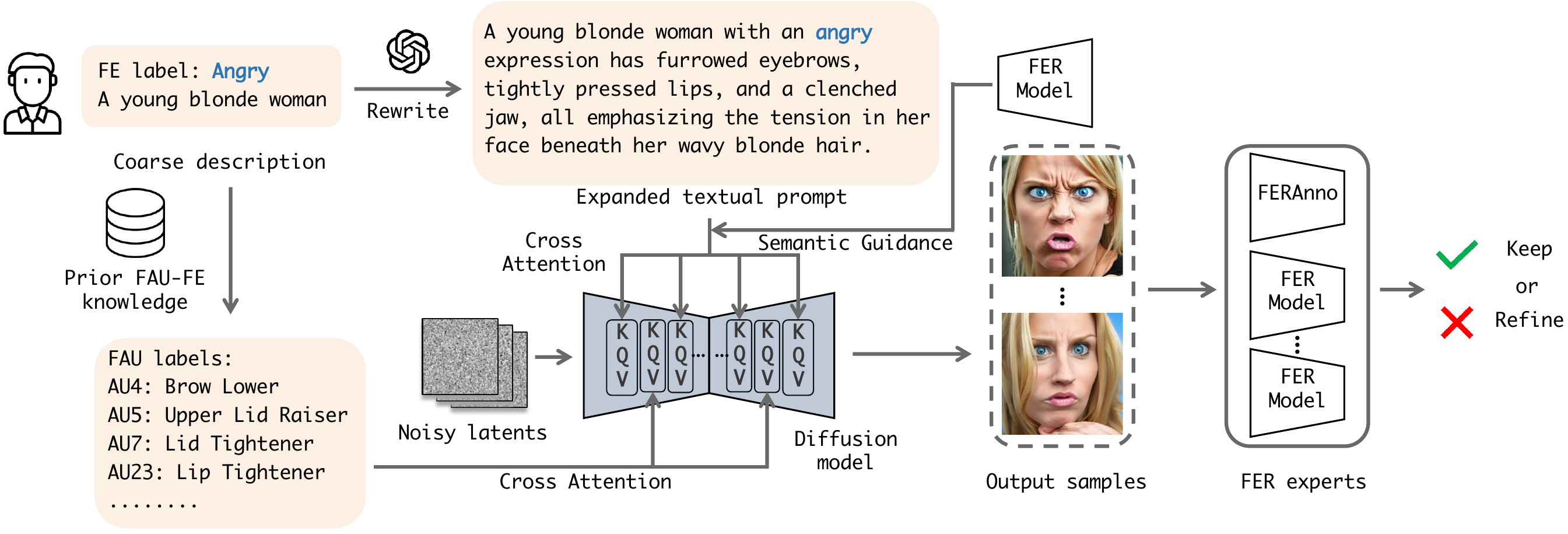}
    \vspace{-0.15in}
    \caption{Overall pipeline of our FER data synthesis process.}
    \label{fig:synpipeline}
    \vspace{-0.2in}
\end{figure*}

\subsection{Overall Pipeline for FER Data Synthesis}


We introduce the overall pipeline for FER data synthesis (Fig.~\ref{fig:synpipeline}). The process starts with a coarse human portrait description assigned to a specific facial expression. ChatGPT enriches this description with details such as facial appearance, subtle facial muscle movements, and contextual cues. Simultaneously, facial action unit annotations are generated based on prior FAU-FE knowledge \cite{ekman1978facial}, aligning them with emotion categories to serve as explicit control signals for guiding the facial expression image synthesis.
Once the facial expression label, facial action unit labels, and expanded textual prompt are prepared, these inputs condition our diffusion model to generate high-fidelity FER images, guided by semantic guidance to ensure accurate FER semantics.
During the denoising process, FERAnno automatically produces pseudo labels for the generated images. 
To further improve labeling accuracy,  we ensemble our FERAnno with existing FER models, which collaborate to vote on the accuracy of the predefined FER labels. 
In cases where discrepancies arise, the predefined label is replaced by averaging the predictions from the ensemble experts. This mechanism effectively reduces the risk of inconsistent or uncertain annotations, ensuring that the final synthesis data is precise and dependable for downstream applications.

\subsection{Diffusion Model Training for FER Data}

\subsubsection{FEText Data Construction}
\label{sec:fetext}

\begin{figure*}[!t]
  \centering
  \includegraphics[width=\textwidth]{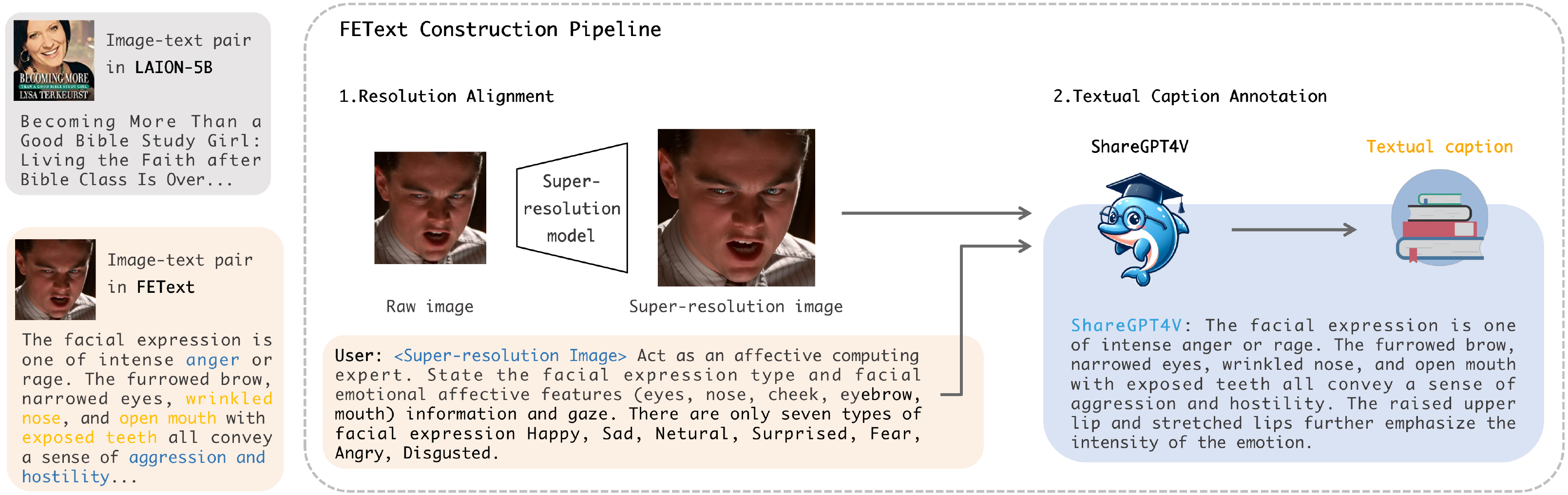}
  \vspace{-2em}
  \caption{Overview of our FEText data construction pipeline.}
  \label{fig:affect_text}
  \vspace{-0.5cm}
\end{figure*}

To address the lack of facial expression image-text pairs for diffusion model training, we introduce FEText (Fig.~\ref{fig:affect_text}), the first hybrid image-text dataset for FER. It combines face images from FFHQ \cite{dataset_ffhq}, CelebA-HQ \cite{dataset_celeba-hq}, AffectNet \cite{dataset_affectnet}, and SFEW \cite{dataset_sfew}, each paired with captions generated by a multi-modal large language model (MLLM). FEText includes 400K curated pairs tailored for facial expression tasks.

\noindent\textbf{Resolution Alignment:} Due to variations in image resolution across different datasets, we first utilize a super-resolution model~\cite{lin2023diffbir} to standardize the resolutions of images from AffectNet and SFEW. Specifically, we incorporate high-resolution images from FFHQ and CelebA-HQ datasets to preserve the model's capacity for high-fidelity image generation. This dual approach allows the model to not only maintain the fidelity of the generated images but also to learn and incorporate the facial expression semantics from AffectNet and SFEW.


\noindent\textbf{Textual Caption Annotation:} To generate a textural caption for each face image, we employ the open-source multi-modal language model ShareGPT-4V~\cite{chen2023sharegpt4v}, by guiding it with carefully crafted instructions. To ensure that the generated captions are both context-aware and expressive, we clearly define the model's role and provide examples of detailed facial expression descriptions within the prompts. This approach enables the model to generate precise, emotion-reflective captions for the input images.

With the FEText obtained, we then conduct the fine-tuning of the diffusion model on it using the diffusion loss in Eq. \ref{eq::diff_loss}. A detailed fine-tuning strategy of the model can be found in the supplementary material.


\subsubsection{Explicit Control Signals via Facial Action Units}
\label{sec:au}

While fine-tuning the diffusion model using facial expression captions provides general language-based guidance for facial expression generation, it lacks the precision needed to capture fine-grained facial details, such as localized muscle movements. To this end, we propose to incorporate more explicit control signals through Facial Action Units (FAUs), each of which represents a specific facial muscle movement.
Inspired by IP-Adapter \cite{ye2023ip}, we apply a decoupled cross-attention module to integrate FAU embeddings with the diffusion model's generation process. These embeddings are derived by mapping discrete FAU labels into high-dimensional representations using a Multi-Layer Perceptron, referred to as the AU adapter. FAU labels for each image in the FEText dataset are annotated using the widely adopted FAU detection model, OpenGraphAU~\cite{luo2022learning}.
With the diffusion model’s parameters frozen, we train the AU adapter to guide the model in recovering facial images based on the annotated FAU labels, using the objective in Eq.~\ref{eq::diff_loss}. 


\subsection{Semantic Guidance for Precise Expression Control}
\label{sec:sg}

Due to the imbalanced distribution of FER labels in the training data and the potential ambiguity between certain facial expressions \cite{zhang2024leave}, such as \textit{disgust}, relying solely on textual and FAU conditions might not guarantee the faithful generation of these expressions.
To address this issue, we propose incorporating semantic guidance on the textual embeddings $c^{\text{text}}$, during the later stages of the denoising process. We leverage external knowledge from open-source FER models to steer the generation process, ensuring a more accurate and faithful synthesis of hard-to-distinguish facial expressions.

\noindent\textbf{Layout Initialization:} 
During inference, we select a random face image $x^{s}$ from FEText and invert it to initialize the noise sample $x_{T}^{s}$(Eq. \ref{eq::add_noise}). Since early diffusion stages shape the global layout of the image \cite{zhang2023inversion, Pan_2023_ICCV, mao2023guided}, this strategy helps preserve the natural facial structure, ensuring the generated images are coherent, high-quality, and visually consistent with real-world expressions.

\noindent\textbf{Semantic Guidance:} In the early steps of the diffusion process, the generation process is conditioned on the original textual condition $c^{\text{text}}$. To further induce the generation of facial expression images corresponding to their FER labels $y$, we iteratively update the textual condition in the subsequent time steps. Specifically, a facial expression classifier $f(\cdot)$ is utilized for the injection of complex semantics. 
To guide the generated images towards the specific class $y$, we propose to do so by updating the textual embeddings. 
Given an intermediate denoised sample $x_t$ at timestep $t$, following Eq. 15 in DDPM \cite{ho2020denoising}, we first estimate the one-step prediction of the original image $\hat{x}_0$ as:
\begin{equation}
    \hat{x}_0 = (x_t - \sqrt{1-\Bar{\alpha}_t})\epsilon_{\theta}(x_t, t, c^{\text{text}}, c^{\text{au}}) / \sqrt{\Bar{\alpha}_t}
\end{equation}
We then calculate the classification loss with:
\begin{equation}
    \mathcal{L}_{g} = -y\log(h(f(\hat{x}_0))_i)
\end{equation}
Given the guidance loss $\mathcal{L}_{g}$, the textual embedding is updated with the corresponding gradient:
\begin{equation}
    c^{\text{text}}_{t-1} = c^{\text{text}}_{t} + \lambda_{g} \frac{ \nabla_{c^{\text{text}}_{t}} \mathcal{L}_{g}} {||\nabla_{c^{\text{text}}_{t}} \mathcal{L}_{g}||_2}
\end{equation}
where $\lambda_g$ and $c_{t-1}^{\text{text}}$ denote the step size and the updated textual embedding at timestep $t-1$, respectively. 
In the latter steps of the diffusion process, the noise estimator network $\epsilon_{\theta}$ is conditioned on the updated textual embeddings rather than the original one.

\subsection{Diffusion-based Label Calibrator (FERAnno)}
\label{sec:anno}

\begin{figure}[!t]
  \centering
  \includegraphics[width=\linewidth]{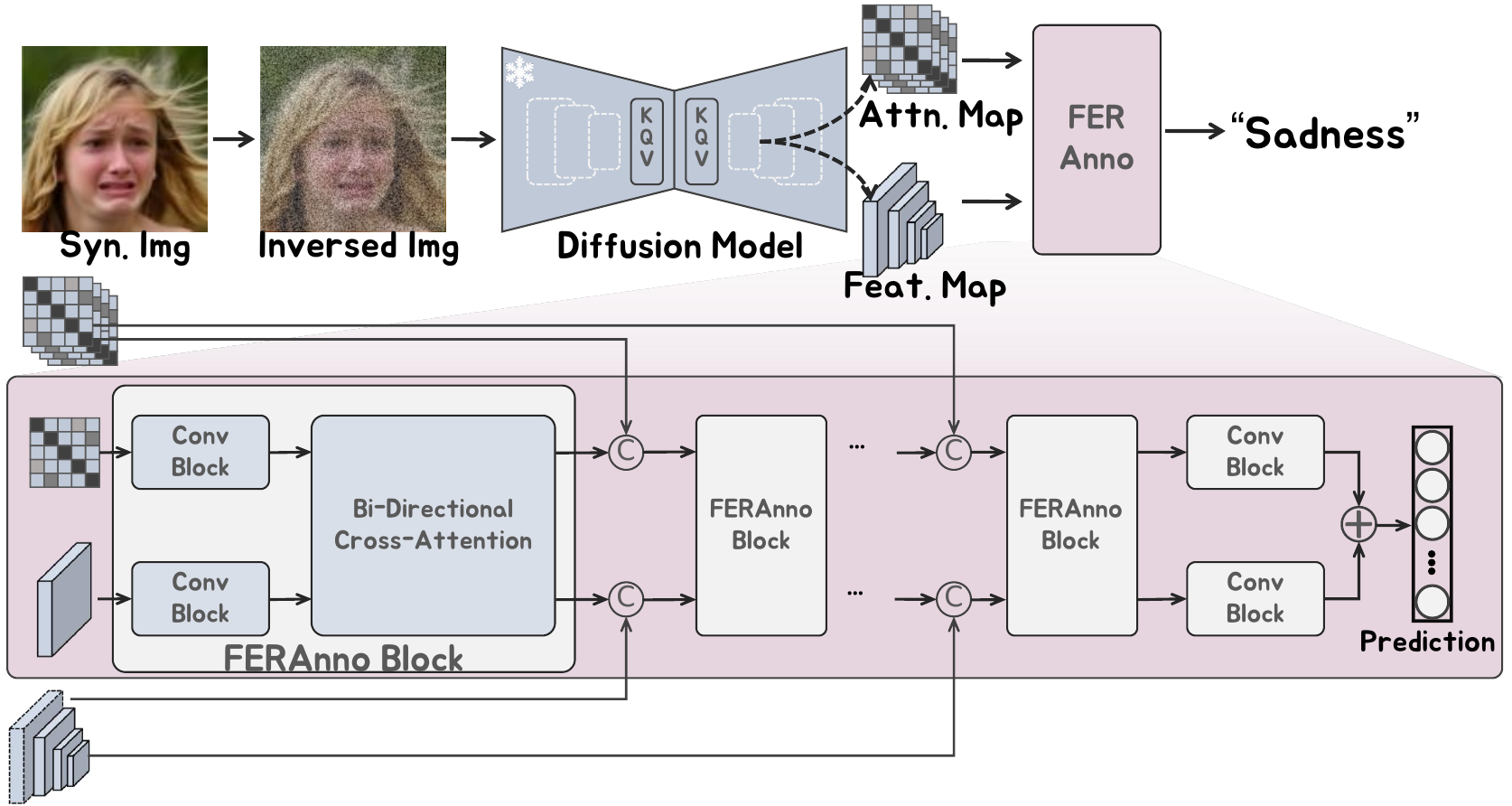}
  \vspace{-0.2in}
  \caption{Overview of our FERAnno pseudo-label generator.}
  \label{fig:feranno}
  \vspace{-0.21in}
\end{figure}

To ensure semantic alignment between each synthesized face image and its assigned facial expression label, we introduce FERAnno, a label calibration framework designed to validate the consistency of the generated data. By analyzing the facial patterns of each synthesized image, FERAnno categorizes them and compares the post-categorized labels with their pre-assigned facial expression labels. This verification process helps identify and filter out samples with mismatched labels, preventing them from negatively impacting downstream FER model training.
Specifically, FERAnno is a diffusion-based label calibrator equipped with a deep understanding of facial semantics. It leverages the multi-scale intermediate features and cross-attention maps inherent in the diffusion model to predict accurate FER labels, as depicted in Fig.~\ref{fig:feranno}. This ensures only high-quality, correctly labeled samples are included in the training pipeline, leading to more reliable model performance.

\noindent\textbf{Image Inversion:} To extract facial features and cross-attention maps with the diffusion model $\epsilon_{\theta}$, we first inverse the generated image $x_0$ back to the noise sample $x_t$ at a denoising timestep $t$, following a predefined scheduler, as described in Eq. \ref{eq::add_noise}. 
To preserve facial details, we set $t = 1$ during the inversion process, ensuring that the facial features remain as close as possible to the original generated image $x_0$. This partially denoised sample is then passed through the trained denoising network, allowing us to extract rich facial features and cross-attention maps from intermediate layers, which are critical for capturing detailed facial patterns.

\noindent\textbf{Feature Extraction:} Given the inverted noise sample $x_1$ and the corresponding textual condition $c^{\text{text}}$ and AU condition $c^{\text{au}}$, we can extract the multi-scale feature representations and textual cross-attention maps from the U-Net $\epsilon_{\theta}$ as $\left\{ \mathcal{F}, \mathcal{A} \right\} = \epsilon_{\theta}(x_1, t_1, c^{\text{text}}, c^{\text{au}})$,
where $\mathcal{F}$ and $\mathcal{A}$ denote the multi-scale feature representations and the cross-attention maps, respectively. $\mathcal{F}$ contains multi-scale feature maps from different layers of the U-Net $\epsilon_{\theta}$ with four different resolutions. $\mathcal{A}$ contains the cross-attention maps drawn from the 16 cross-attention blocks in $\epsilon_{\theta}$. Both the feature representation $\mathcal{F}$ and the cross-attention maps $\mathcal{A}$ are regrouped according to their resolutions.

\noindent\textbf{Multi-scale Features and Attention Maps Fusion:}
Given that the multi-scale feature maps $\mathcal{F}$ capture global information essential for image generation, and the cross-attention maps provide class-discriminative information as well as relationships between object locations \cite{tang2022daam, caron2021emerging}, FERAnno fuses both features and attention maps within a dual-branch encoder architecture for pseudo-label annotation. An overview of this architecture is shown in Fig. \ref{fig:feranno}.
We first compute the mean of the regrouped attention maps, denoted as $\mathcal{A}_{\text{reg}}$, yielding a set of averaged attention maps $\Bar{\mathcal{A}}$. Both the feature maps $\mathcal{F}$ and the averaged attention maps $\Bar{\mathcal{A}}$ are then passed through a residual convolution block to prepare them for further processing. To effectively integrate information at different scales, we introduce a bi-directional cross-attention block to fuse the features and attention maps. 1$\times$1 convolutions are employed at various stages to adapt the fusion across multiple resolution layers. Finally, the fused feature maps and attention maps are concatenated and passed through a linear layer, which outputs a probability vector for predicting facial expression classes.
\section{Experiments}
\vspace{-0.5em}
We conduct extensive experiments to evaluate both the generation quality of our synthetic data (Sec.~\ref{sec:gen_quality}) and its effectiveness in FER tasks (Sec.~\ref{sec:effec_syndata}). 
Implementation details are in the appendix.

\noindent\textbf{Evaluation metrics:} 
In the experimental evaluation, we employ both objective metrics and user studies to comprehensively assess the generation quality. For synthetic image quality, we utilize Fréchet Inception Distance (FID) \cite{yu2021frechet} to measure distribution similarity with real data, FaceScore (FS) \cite{Liao2024FaceScoreBA} for facial quality assessment, and Human Preference Score v2 (HPSv2) \cite{wu2023human} with Multi-dimensional Preference Score (MPS) \cite{zhang2024learning} for human perception evaluation. Facial expression accuracy is quantified using pre-trained classifiers and Facial Action Unit Accuracy (FAU Acc.) via AU detection models. For subjective user study, the subjects are asked to select the images with better expression alignment and face fidelity for pairs of synthetic images generated by SynFER and other baselines.
Downstream task performance is evaluated through linear probing accuracy in self-supervised learning, classification accuracy in supervised settings, and both Weighted Average Recall (WAR) and Unweighted Average Recall (UAR) for zero-shot recognition. Few-shot learning capabilities are measured using standard n-way k-shot protocols across compound expression datasets.



\subsection{Generation Quality}
\label{sec:gen_quality}

\begin{table*}[htbp]
\centering
\setlength{\tabcolsep}{20pt}
\resizebox{0.95\linewidth}{!}{
\begin{tabular}{lcccccccc}
\midrule
\multirow{2}{*}{\textbf{Method}} & \multicolumn{6}{c}{\textbf{Objective Metrics}} & \multicolumn{2}{c}{\textbf{User study (Ours vs. )(\%)}}\\
\cmidrule(r){2-7}\cmidrule(r){8-9} & FID ($\downarrow$) &HPSv2($\uparrow$) & FS($\uparrow$)  & MPS ($\downarrow$)   
& FER Acc.($\uparrow$)   &   FAU Acc.($\uparrow$)
& EA ($\uparrow$) & FF ($\uparrow$)\\ 
\midrule\midrule
Stable Diffusion    & 88.40  &     0.263 & 2.01 &   2.00    &   20.06   &   87.72 & 2.86 & 1.79\\
PixelArt    & 145.23  &      0.271 & 3.79   &   5.26    &   15.52   &   84.57 & 24.26 & 10.00\\
PlayGround  & 81.76  &      0.265 & 2.86    &   3.73    &   21.56   &   87.28 & 7.50 & 5.00\\
FineFace    & 74.61  &      0.268    &  3.29 & 1.48    &   38.05   &   89.68 & 5.73 & 6.41\\
\midrule
\textbf{SynFER}  & \textbf{16.32} &  \textbf{0.280}    &  \textbf{4.26} & \textbf{0.50}   &   \textbf{55.14}   &   \textbf{93.31} & \textbf{59.64} & \textbf{76.79}\\
\midrule
\end{tabular}
}
\vspace{-1em}
\caption{Generation quality comparisons. Ours vs.' shows the proportion of users who prefer our method over the alternative. An MPS above 1.00 and results above 50\% in the user study indicate SynFER outplays the counterpart. FS, FER Acc., FAU Acc., EA and FF denote FaceScore \cite{Liao2024FaceScoreBA}, FER accuracy, facial action unit accuracy, expression alignment and face fidelity, respectively.
}
\label{tab:gen_quality}
\vspace{-0.1in}
\end{table*}

We present both objective metrics and subjective user studies, comparing our method to state-of-the-art (SOTA) diffusion models \cite{ldm, chen2023pixart, li2024playground} and the latest facial expression generation technique, FineFace \cite{fineface}.
We compute FID between the synthesis images and the test set of the AffectNet \cite{dataset_affectnet}.
Tab. \ref{tab:gen_quality} shows that our method outperforms popular diffusion models and SOTA facial expression generation method FineFace \cite{fineface}, across all metrics of image quality, human preference and facial expression accuracy. Notably, the advantages of SynFER in both FE Acc. and AU Acc. indicate its outstanding controllability in facial expression generation.

\subsection{Effectiveness of Synthetic Dataset}
\label{sec:effec_syndata}

\begin{table}[]
    \centering
\centering
    \setlength{\tabcolsep}{2pt}
    \scalebox{0.66}{ 
        \begin{tabular}{l|p{2.5cm}p{1.5cm}p{2cm}p{2cm}p{2cm}}
\toprule
\multirow{2}{*}{\textbf{Method}} & \multicolumn{2}{c}{\textbf{Pre-train Data}} & \multirow{2}{*}{\textbf{RAF-DB}} & \multirow{2}{*}{\textbf{AffectNet}} & \multirow{2}{*}{\textbf{SFEW}} \\
\cmidrule{2-3}
~ & \textbf{Dataset} & \textbf{Scale} & \\ \midrule\midrule
MCF & Laion-Face & 20M & 65.22 & -& 32.61 \\
FRA & VGGFace2 & 3.3M & 73.89 & 57.38 & - \\
PCL & VoxCeleb & 1.8M & 74.47 & - & 39.68 \\
\midrule
SimCLR & AffectNet & 0.2M & 78.65 & 48.36 & 46.79 \\ 
SimCLR & Ours & 1.0M & 80.24 \textcolor{red}{(+1.59)} & 52.05 \textcolor{red}{(+3.69)}& 47.62 \textcolor{red}{(+0.83)} \\
SimCLR & AffectNet+Ours & 1.2M & 81.52 \textcolor{blue}{(+2.87)} & 54.37 \textcolor{blue}{(+6.01)}& 48.52 \textcolor{blue}{(+1.73)} \\
\midrule
BYOL & AffectNet & 0.2M & 78.24 & 50.04& 48.70 \\ 
BYOL & Ours & 1.0M & 80.96 \textcolor{red}{(+2.72)} & 53.13 \textcolor{red}{(+3.09)}  & 51.35 \textcolor{red}{(+2.65)} \\  
BYOL & AffectNet+Ours & 1.2M & 81.25 \textcolor{blue}{(+3.01)} & 54.95 \textcolor{blue}{(+4.91)}  & \textbf{51.70} \textcolor{blue}{(+3.00)} \\
\midrule
MoCo v3 & AffectNet & 0.2M & 79.05 & 51.03& 49.34 \\ 
MoCo v3 & Ours & 1.0M & 81.17 \textcolor{red}{(+2.12)} &  55.56\textcolor{red}{(+4.53)} & 50.78 \textcolor{red}{(+1.44)} \\ 
MoCo v3 & AffectNet+Ours & 1.2M & \textbf{81.68} \textcolor{blue}{(+2.63)} & \textbf{57.84} \textcolor{blue}{(+6.81)} & 51.26 \textcolor{blue}{(+1.92)} \\
\bottomrule
\end{tabular}
    }
    \vspace{-1em}
    \caption{Linear probe performance comparisons of SSL models on three FER datasets.}
    \label{tab:fer_performance}
    \vspace{-1em}
\end{table}

\begin{table}[]
    \centering
\setlength{\tabcolsep}{25pt}
    \centering
    \scalebox{0.75}{ 
            \begin{tabular}{l|c|c}
            \toprule
            \textbf{Method} & \textbf{RAF-DB} & \textbf{AffectNet} \\
            \midrule\midrule
            ResNet-18 & 87.48 & 50.32 \\ 
            ResNet-18 + Ours & 87.97 & 51.65 \\
            Ada-DF & 90.94 & 65.34 \\ 
            Ada-DF + Ours & 91.21 & 66.82 \\
            POSTER++ & 91.59 & 67.49 \\ 
            POSTER++ + Ours & 91.95 & 69.04 \\
            APViT & 91.78 & 66.94 \\ 
            APViT + Ours & 92.05 & 67.26 \\
            \midrule 
            FERAnno & \textbf{92.56} & \textbf{70.38} \\
            \bottomrule
            \end{tabular}
    }
    \vspace{-1em}
    \caption{Comparison of supervised learning models (with and without our synthetic data) and the label calibrator FERAnno.}
    \label{tab:sup} \vspace{-1em}
\end{table}


\noindent\textbf{Self-supervised Representation Learning:}
We trained self-supervised learning (SSL) models, including BYOL \cite{grill2020bootstrap}, MoCo v3 \cite{chen2021empirical}, and SimCLR \cite{chen2020simple}, using real-world data, our synthetic data, and a combination of both. The linear probe performances are evaluated on three widely used facial expression recognition (FER) datasets: RAF-DB \cite{dataset_rafdb}, AffectNet \cite{dataset_affectnet}, and SFEW \cite{dataset_sfew}, with results reported in Tab.~\ref{tab:fer_performance}. All SSL models are trained with a ResNet-50 architecture~\cite{he2016deep}.
Notably, SOTA self-supervised facial representation learning, such as MCF \cite{ssl_mcf}, FRA \cite{ssl_fra}, and PCL \cite{ssl_pcl}, are pre-trained on much larger face datasets like LAION-Face~\cite{laion-face}, VGGFace2~\cite{cao2018vggface2}, and VoxCeleb~\cite{nagrani2020voxceleb}. However, these models underperformed on FER tasks compared to ours, highlighting that existing large-scale face datasets may lack the high-quality and diverse facial expression patterns required for accurate FER.
Results show that combining real-world and synthetic data consistently boosts SSL baselines. Remarkably, even when MoCo v3 was trained solely on our synthetic data, it achieved a 2.12\% improvement on RAF-DB, underscoring the effectiveness of our approach in capturing critical facial expression details that are essential for FER.

\noindent\textbf{Supervised Representation Learning:}
We validate the effectiveness of SynFER for supervised representation learning by evaluating its performance on RAF-DB and AffectNet (Tab.~\ref{tab:sup}). 
We compare with SOTA FER models, including Ada-DF \cite{liu2023dual}, POSTER++ \cite{mao2024poster++}, and APViT \cite{xue2022vision}. The results demonstrate that incorporating synthetic data consistently enhances both baseline models and the latest SOTAs in supervised facial expression recognition. Notably, APViT benefits from the synthetic data with improvements of 0.27\% on RAF-DB and 0.32\% on AffectNet.
While the improvements in supervised learning are more modest compared to self-supervised learning, they remain consistent. This is likely due to the stricter distribution alignment required in supervised learning between synthetic training data and real-world test data. In the following section on scaling behavior analysis, we provide further insights, showcasing the use of the distribution alignment technique, Real-Fake \cite{real-fake}, to alleviate this problem.

\begin{table}[!t]
    \centering
    \setlength{\tabcolsep}{7pt}
    \scalebox{0.75}{
        \begin{tabular}{c|cc|cc|cc}
            \toprule
            \multirow{2}{*}{\textbf{Method}} & \multicolumn{2}{c|}{\textbf{CFEE\_C}} & \multicolumn{2}{c|}{\textbf{EmotionNet\_C}} & \multicolumn{2}{c}{\textbf{RAF\_C}} \\
            & \textbf{1-shot} & \textbf{5-shot} & \textbf{1-shot} & \textbf{5-shot} & \textbf{1-shot} & \textbf{5-shot} 
            \\\midrule\midrule
            InfoPatch & 54.19 & 67.29 & 48.14 & 59.84 & 41.02 & 57.98 \\ 
            InfoPatch* & 55.21 & 68.73 & 48.52 & 61.16 & 41.88 & 59.54 \\\midrule
            LR+DC & 53.20 & 64.18 & 52.09 & 60.12 & 42.90 & 56.74\\ 
            LR+DC* & 54.65 & 65.28 & 51.96 & 60.14 & 43.87 & 57.90\\\midrule
            STARTUP & 54.89 & 67.79 & 52.61 & 61.95 & 43.97 & 59.14\\ 
            STARTUP* & 56.25 & 69.93 & 52.87 & 62.12 & 45.18 & 61.23\\\midrule
            CDNet & 56.99 & 68.98 & 55.16 & 63.03 & 46.07 & 63.03\\ 
            CDNet* & \textbf{57.74} & \textbf{70.64} & \textbf{56.79} & \textbf{65.63} & \textbf{46.97} & \textbf{64.34}\\
            \bottomrule
            \end{tabular}
    }
    \vspace{-1em}
    \caption{Comparisons with SOTA few-shot learning methods on 5-way few-shot FER tasks with a 95\% confidence interval. (*) indicates training with both real-world and synthetic data.
    }
    \label{tab:few-shot}
    \vspace{-1em}
\end{table}

\noindent\textbf{Few-shot Learning:} We explore the potential of synthetic data to enhance few-shot learning, as presented in Tab.~\ref{tab:few-shot}. Following the protocol from CDNet \cite{zou2022learn}, we train models on five basic expression datasets and evaluate on three compound expression datasets: CFEE\_C \cite{cfee}, EmotionNet\_C \cite{emotionet}, and RAF\_C \cite{dataset_rafdb}. We compare it against SOTA few-shot learning methods, including InfoPatch \cite{liu2021learning}, LR+DC \cite{DBLP:conf/iclr/YangLX21}, and STARTUP \cite{DBLP:conf/iclr/PhooH21}.
The results clearly demonstrate that integrating synthetic data consistently enhances few-shot FER performance across key metrics. This highlights the potential of synthetic data in data-limited scenarios, allowing models to better generalize to complex, real-world expressions in few-shot tasks.

\vspace{-1em}

\begin{figure}
    \centering
    \setlength{\tabcolsep}{2pt}
    \includegraphics[width=0.9\linewidth]{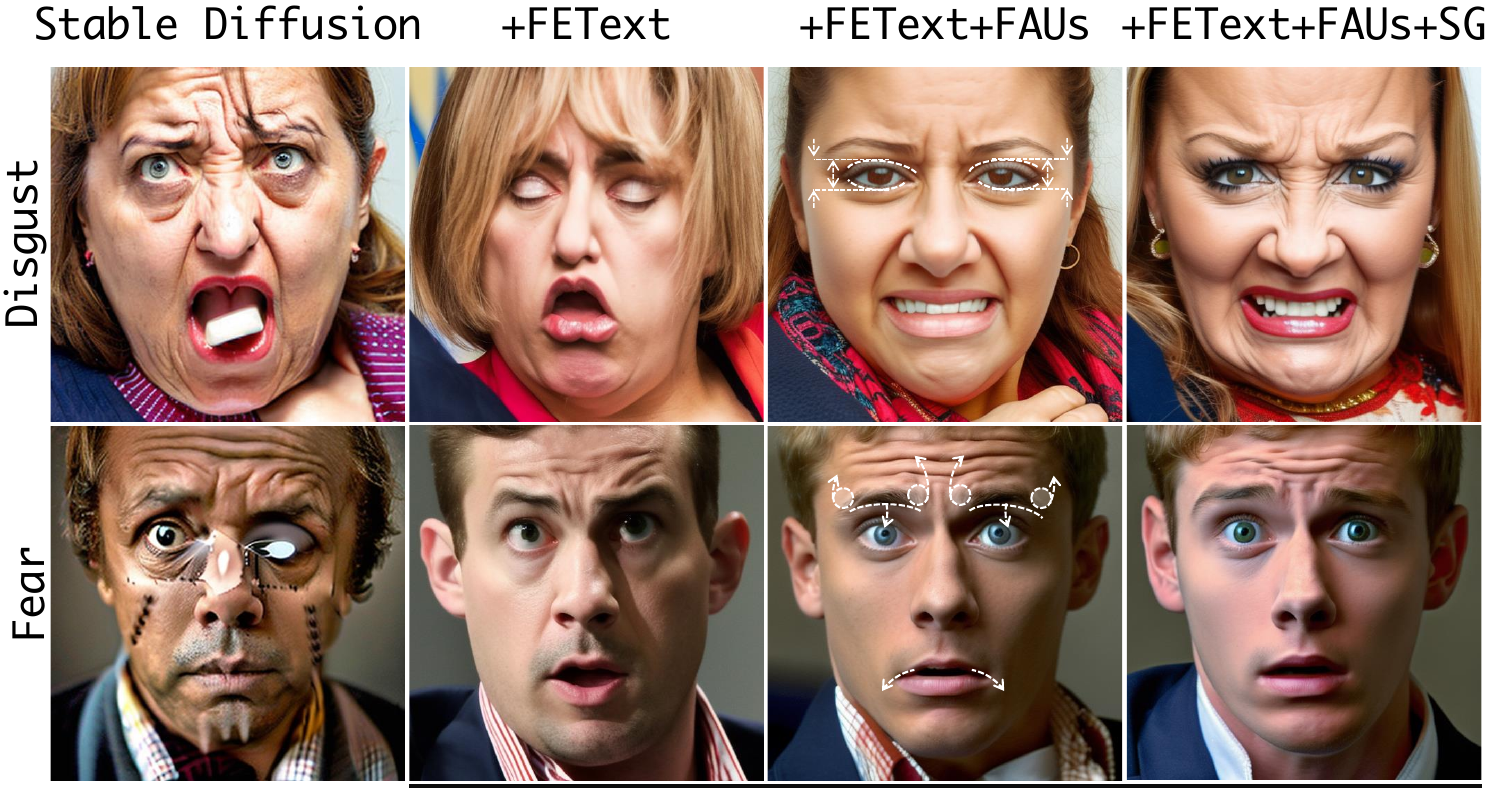}
    \vspace{-1em}
    \caption{Example of generated samples.}
    \label{fig:gen_vis} \vspace{-2.6em}
\end{figure}

\begin{figure*}[!ht]
    \centering
    \subfloat[Synthetic Data Scaling in Self-Supervised Learning]{\includegraphics[width=0.3\textwidth]{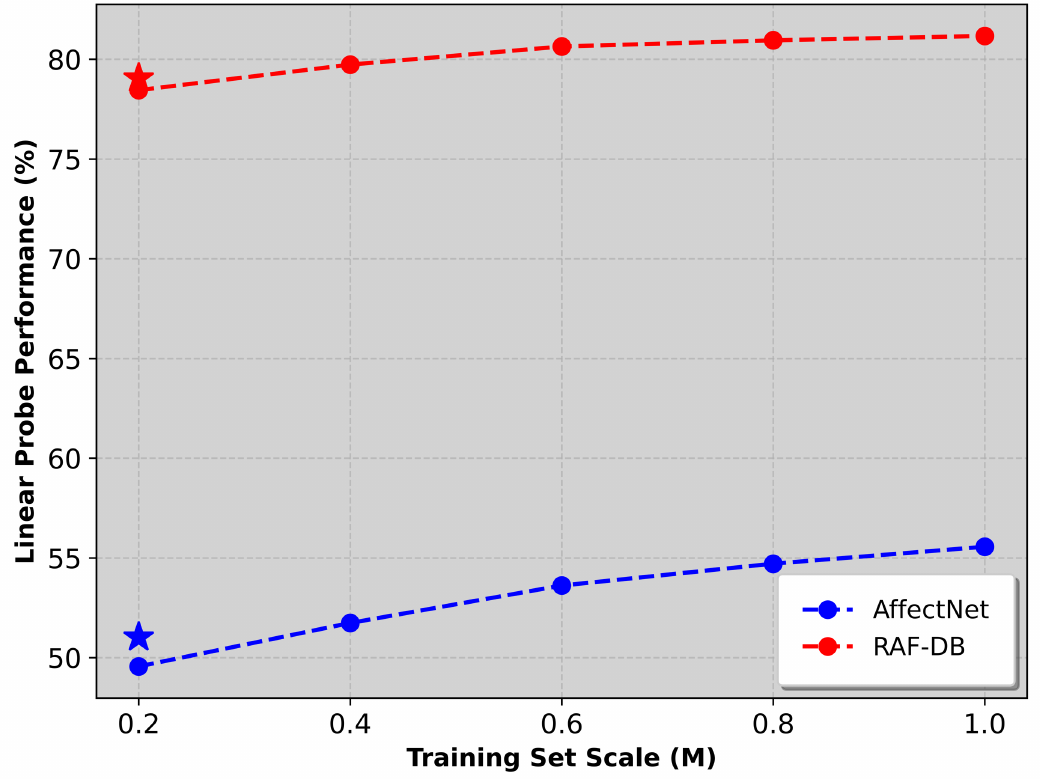}}
    \hspace{2mm}
    \subfloat[Synthetic Data Scaling in Supervised Learning]{\includegraphics[width=0.3\textwidth]{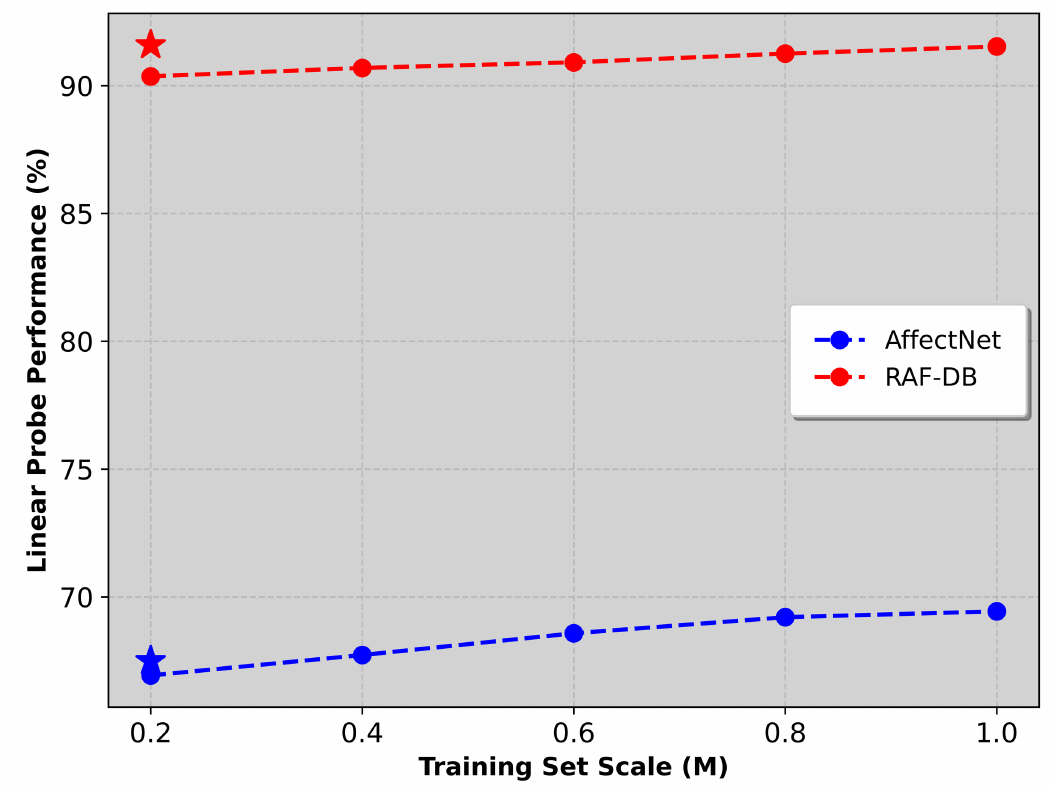}}
    \hspace{2mm}
    \subfloat[Synthetic Data Scaling in Supervised Learning (with Real-Fake technique~\cite{real-fake})]{\includegraphics[width=0.3\textwidth]{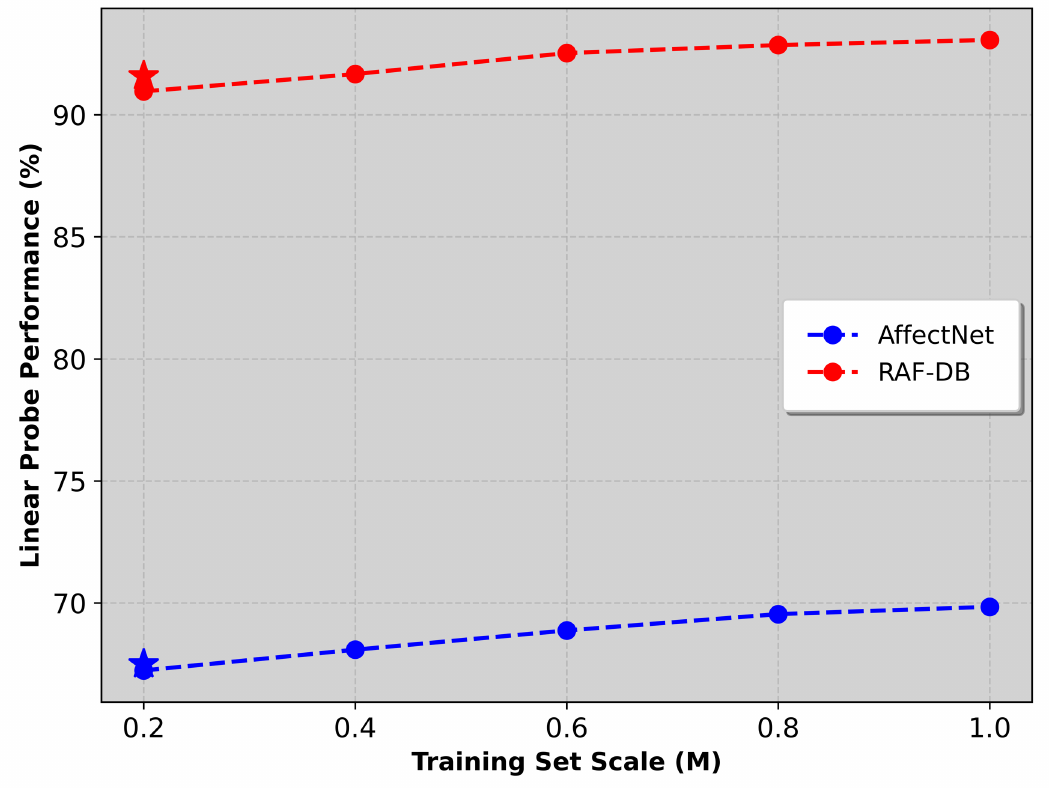}}
    \vspace{-1em}
    \caption{Scaling up the synthetic dataset with MoCo v3 (ResNet-50) \cite{chen2021empirical}, and linear probe performance is evaluated on AffectNet and RAF-DB. The SOTA FER model, POSTER++ \cite{mao2024poster++}, is trained using supervised learning (with and without the Real-Fake technique \cite{real-fake}) on our synthetic dataset and evaluated on the same two target FER datasets. \ding{72} is model's performance trained on corresponding real data.}
    \label{fig:scale}
    \vspace{-1.5em}
\end{figure*}

\subsection{Ablation Study}

\textbf{Effectiveness of FAU Control.}  
Fig.~\ref{fig:gen_vis} shows samples generated with FAU control (third column) exhibit facial expressions that more accurately match their assigned labels compared to those generated with only text guidance (second column). For example, the 'fear' expression, driven by FAUs like Inner Brow Raiser and Lip Stretcher, becomes more distinct (third column, second row), making it easier to differentiate from other emotions such as 'surprise.' Similarly, 'disgust' is more pronounced with FAUs like Lid Tightener. Without FAU control, facial expressions (second column) tend to blur, as different categories show overlapping features.
Quantitative results in Tab.~\ref{tab:ausg} highlight the impact of FAU control: FER accuracy increases from 34.62\% to 48.74\%, and FAU detection accuracy rises from 88.91\% to 92.37\%. This also translates into improved downstream performance on RAF-DB and AffectNet.

\begin{table}[]
    \centering
    \setlength{\tabcolsep}{2pt}
    \resizebox{\linewidth}{!}{
    \begin{tabular}{c|ccc|cc}
    \toprule
    \textbf{Method}  & \textbf{HPSv2} & \textbf{FE Acc.} & \textbf{AU Acc.} & \textbf{RAF-DB} & \textbf{AffectNet} \\ \midrule\midrule
    Real-world Data & - & - & - & 91.59 & 67.49 \\
    SD & 0.263 & 20.06 & 87.72 & 89.42 & 65.36\\ \midrule
    w/ FEText & 0.267 & 34.62 & 88.91 & 90.54 & 66.62\\
    w/ FEText+FAUs & 0.275 & 48.74 & 92.37 & 91.68 & 67.68\\
    w/ FEText+FAUs+SG & \textbf{0.280} & \textbf{55.14} & \textbf{93.31} & \textbf{91.95} & \textbf{68.13}\\
    \bottomrule
    \end{tabular}
    }
    \vspace{-1em}
    \caption{Ablation study on AU injection and semantic guidance (SG) on both generation quality and supervised learning. SD denotes Stable Diffusion \cite{rombach2022high}, which is used as a baseline.}
    \label{tab:ausg}
    \vspace{-1.5em}

\end{table}

\noindent\textbf{Effectiveness of Semantic Guidance:} We further explore the impact of semantic guidance (SG) on both generation quality and supervised representation learning ( Fig.~\ref{fig:gen_vis} and Tab.~\ref{tab:ausg}). By updating text embeddings to better align with the target facial expression category, SG improves the accuracy of the generated expressions by 6.4\%, compared to static text and FAUs. The samples in the last column of Fig.~\ref{fig:gen_vis} show more exaggerated facial expressions than those in the third column, with SG enhancing the intensity.

\begin{figure}[!h]
  \centering
  \vspace{-0.6em}
  \includegraphics[width=0.8\linewidth]{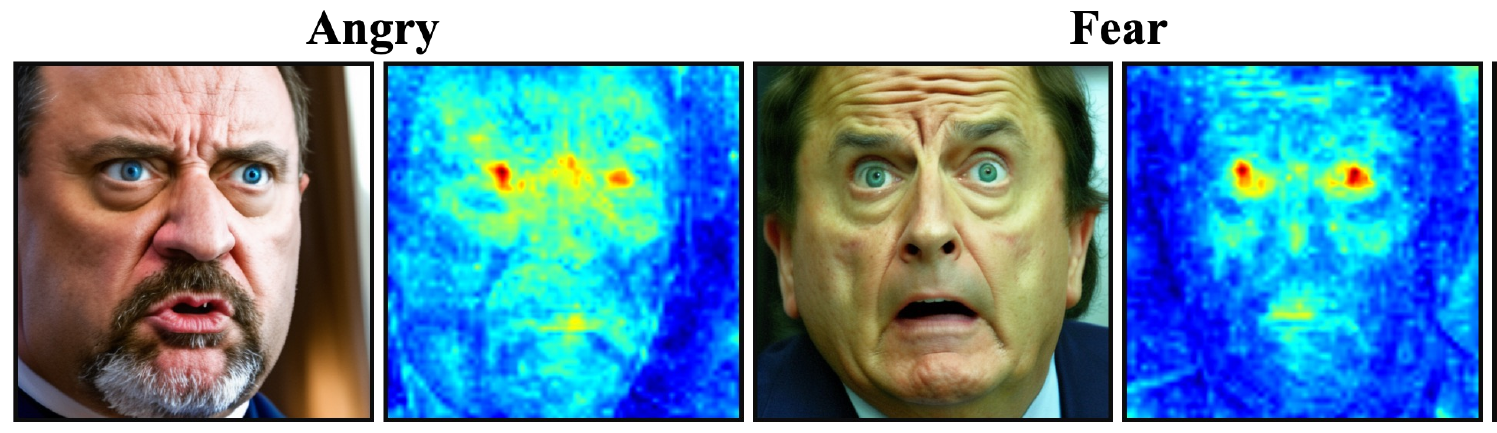}
  \vspace{-1em}
  \caption{Synthesis images and attention maps in the fine-tuned diffusion model.}
  \label{fig:attnmap}
  \vspace{-1.8em}
\end{figure}

\noindent\textbf{Reliability of FERAnno:} We assess the reliability of FERAnno as a label calibrator by evaluating its performance on two FER datasets and visualizing its attention maps in Tab.~\ref{tab:sup} and Fig.~\ref{fig:attnmap}. 
FERAnno outperforms FER SOTAs by +0.51\% and +1.34\% on RAF-DB and AffectNet over the second-best models.
The attention maps (Fig.~\ref{fig:attnmap}) further show FERAnno’s ability to accurately locate facial expression-related patterns e.g., jaw-dropping and furrowed eyebrows, highlighting its semantic understanding.




\noindent\textbf{Synthetic Data Scaling Analysis:} Following \cite{tian2024stablerep, Fan_2024_CVPR}, we investigate the scaling behavior of synthetic data in both self-supervised and supervised learning paradigms. 
We train models exclusively on synthetic images, without combining real-world data. The results in Fig.~\ref{fig:scale} (a)-(b) show a stronger scaling effect in self-supervised learning compared to supervised learning, where performance improves significantly with more data. (c). Compared to standard supervised learning, Real-Fake demonstrates a clear performance boost, which is likely due to the need for better distribution alignment in supervised learning \cite{real-fake}. While SynFER focuses on addressing FER data scarcity, aligning the synthetic data distribution with real-world data is crucial for supervised tasks. To further explore this, we apply the Real-Fake technique \cite{real-fake} for real and synthetic data distribution alignment, and present the results in Fig.~\ref{fig:scale} (c). Compared to standard supervised learning, Real-Fake demonstrates a clear performance boost.



\section{Conclusion}


We propose a synthetic data framework SynFER for facial expression recognition to address the data shortage in the field. 
We introduce the first facial expression-related image-text pair dataset FEText. We inject facial action unit and external knowledge from existing FER models to ensure fine-grained control and faithful generation of the facial expression images. 
To incorporate the generated images into training, we propose a diffusion-based label calibrator to help rectify the annotations for the synthesized images. After constructing the data synthesis pipeline, we show the effectiveness of synthesis data across different learning paradigms. Limitations of this work are discussed in detail in supplementary material.

\section{Acknowledgments}
The work was supported by the National Natural Science Foundation of China under grants no. 62276170, 82261138629, 62306061, 
the Science and Technology Project of Guangdong Province under grants no. 2023A1515010688, the Science and Technology Innovation Commission of Shenzhen under grant no. JCYJ20220531101412030, Open Research Fund from Guangdong Laboratory of Artificial Intelligence and Digital Economy (SZ) under Grant No. GML-KF-24-11, and Guangdong Provincial Key Laboratory under grant no. 2023B1212060076. 
The research reported in this publication was supported by funding from King Abdullah University of Science and Technology - Center of Excellence for Generative AI, under award number 5940.
{
    \small
    \bibliographystyle{ieeenat_fullname}
    \bibliography{main}
}

\end{document}